\documentclass[letterpaper, 10 pt, conference]{ieeeconf}  
\usepackage{graphicx}
\usepackage{graphicx}
\usepackage[export]{adjustbox}
\usepackage{setspace}
\usepackage{subfigure}
\usepackage{mathptmx}
\usepackage{bm}
\usepackage{amsmath,amssymb} 
\usepackage{color}
\usepackage{algorithm}
\usepackage{algpseudocode}
\algnewcommand{\LineComment}[1]{\State \(\triangleright\) #1}
\usepackage{enumerate}
\usepackage{epstopdf}
\usepackage{array}  
\usepackage{eqparbox}
\usepackage{multirow}
\usepackage{url}
\usepackage{mathtools}
\usepackage{hyperref}
\usepackage{adjustbox}
\usepackage{caption}
\usepackage{textcomp}

\usepackage[T1]{fontenc}
\usepackage[font=small,labelfont=bf,tableposition=top]{caption}

\usepackage{array}
\usepackage{mathtools,eqparbox}
\usepackage{tikz}
\usepackage{lipsum}
\usepackage{float}

\usepackage{amsmath} 
\usepackage{subfigure}
\usepackage{epstopdf}
\usepackage{url}
\usepackage{multirow}
\usepackage{algorithm}
\usepackage{algorithmicx}
\usepackage{media9}
\usepackage{multimedia}
\usepackage{subfigure}
\usepackage{algpseudocode}
\usepackage{amsmath} 
\usepackage{amssymb}
\usepackage{graphicx}

\IEEEoverridecommandlockouts                              

\title{\LARGE 
\textbf{Predictive Barrier Lyapunov Function Based Control for Safe Trajectory Tracking of an Aerial Manipulator} 
}

\author{Vedant Mundheda$^{1}$, Karan Mirakhor$^{2}$, Rahul K S$^{3}$, Harikumar Kandath$^{1}$ and Nagamanikandan Govindan$^{1}$
\thanks{$^{1}$Robotics Research Center, IIIT-Hyderabad, India {\small vedant.mundheda@research.iiit.ac.in, harikumar.k@iiit.ac.in, nagamanikandan.g@iiit.ac.in}, $^{2}$TCS Research, Kolkata, India {\small karan.mirakhor@tcs.com}, $^{3}$University of Michigan {\small rahulswa@umich.edu}}}

\begin{document}

\maketitle
\thispagestyle{empty}
\pagestyle{empty}

\begin{abstract}

This paper proposes a novel controller framework that provides trajectory tracking for an Aerial Manipulator (AM) while ensuring the safe operation of the system under unknown bounded disturbances. The AM considered here is a 2-DOF (degrees-of-freedom) manipulator rigidly attached to a UAV. Our proposed controller structure follows the conventional inner loop PID control for attitude dynamics and an outer loop controller for tracking a reference trajectory. The outer loop control is based on the Model Predictive Control (MPC) with constraints derived using the Barrier Lyapunov Function (BLF) for the safe operation of the AM. BLF-based constraints are proposed for two objectives, viz. 1) To avoid the AM from colliding with static obstacles like a rectangular wall, and 2) To maintain the end effector of the manipulator within the desired workspace.
The proposed BLF ensures that the above-mentioned objectives are satisfied even in the presence of unknown bounded disturbances.
The capabilities of the proposed controller are demonstrated through high-fidelity non-linear simulations with parameters derived from a real laboratory scale AM. We compare the performance of our controller with other state-of-the-art MPC controllers for AM.  

\end{abstract}

\section{INTRODUCTION}

\par
\begin{figure}
    \centering
    \includegraphics[width=0.35\textwidth]{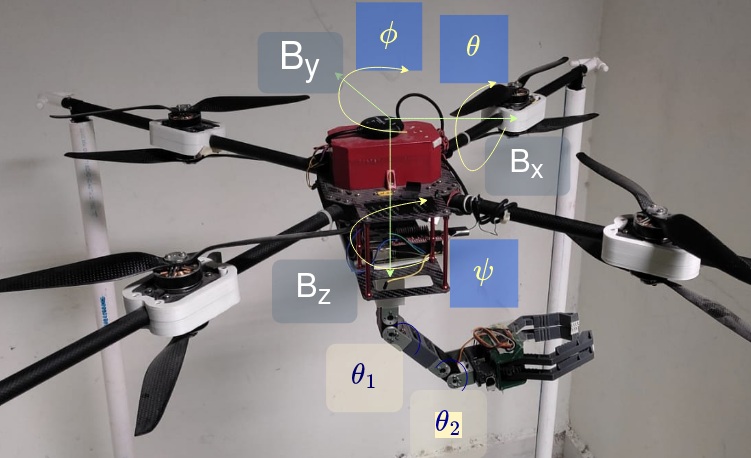}
    \caption{Aerial Manipulator as in RRC, IIIT Hyderabad}
    \label{fig1}
\end{figure}

 Aerial Manipulators (AMs) have gained much attention in recent years \cite{AM literature review}. The Unmanned Aerial Vehicle (UAV) acts as a floating base for the manipulator enabling it to conduct active operations in the 3D space, see Fig. \ref{fig1}. This combination of a UAV and a manipulator provides the system with enough capabilities to perform a  range of complex operations where human access is limited (e.g., in disaster-affected zones). The other industrial and commercial applications of AMs are in the maintenance of power grids, the inspection of bridges, and canopy sampling \cite{AM literature review}.

Such applications involve trajectory tracking maneuvers by the AM near static objects like bridges, trees, and buildings. It is unarguable that any system should be designed to be safe. In fact, safety has been a major hurdle in deploying such AM systems in these applications \cite{Safety issues}.
Three major issues are faced with said close maneuvers. Firstly, hovering close to such objects leads to ground, ceiling, and wall effects, causing immeasurable turbulent disturbances 
\cite{Ground effect}. Secondly, the AM encounters disturbances in the form of forces and torques due to the highly coupled dynamics of the UAV, and manipulator \cite{Robotica: Modeling and control}. These disturbances can lead to instability of the AM and cause a collision with the obstacles. E.g., external factors in the form of wind disturbances can lead to instability \cite{Wind disturbances}. Thirdly, it is only sometimes possible to accurately model the obstacles around a trajectory due to their irregular shape, lack of visibility, or uncertainty associated with obstacle locations. This can occur when the AM maneuvers through a dark or uneven tunnel incapacitating it to determine a bound across the obstacles. 
Keeping this in mind, the AM must operate while keeping a safe distance from obstacles and maintaining stability. The AM movement can be bound in a desired workspace around the desired trajectory. This will prevent any possible collisions with obstacles. 

\noindent \textbf{Related Work} 

\noindent While the literature is sufficiently populated with novel design approaches of AMs \cite{Design 1}-\cite{Design 2}, prior work involving safe control of the AM has been sparse.   
 Adaptive controller\cite{Adaptive control} tackles torques due to the highly coupled dynamics of the AM by using an outer loop adaptive control over the proportional–derivative (PD) inner loop of the UAV. Though it provides computational efficiency, it is incapable of incorporating constraints to avoid obstacles.  
Model Predictive Control (MPC) \cite{NMPC} significantly reduces the abruptness in control inputs and tracks the desired trajectory while anticipating future dynamic interactions of the AM. PID and traditional adaptive controllers lack this predictive ability. MPC is applied to open a hinged door \cite{MPC Door open}. Considering the coupled dynamics of an AM and a hinged door, an MPC in the framework of a Linear Quadratic Regulator is designed.

Barrier Lyapunov Function (BLF) \cite{CBF Intro} is used as a tool to enforce the safety of non-linear dynamical systems. Barrier certificates \cite{ZCBF} are established considering a safe region of operation defined as $\zeta$. While guaranteeing forward invariance of $\zeta$,  safety is ensured.
BLF-based MPC for a non-linear system described in \cite{MPC BLF NL} proposes a stabilizing controller to ensure avoidance of a set of states associated with the unsafe region for a chemical process.  
MPC combined with constraints using BLF \cite{P Mali} is used for distributive multi-UAV avoidance. 
An MPC scheme for safety planning \cite{MPC safety} using BLF demonstrates the trade-off between the safety and performance of a UAV.


\noindent \textbf{Contributions} 

\noindent The key contributions of this paper are the following:

\begin{enumerate}
    \item To the best of the author's knowledge, this is the first attempt to incorporate safe operation for AM maneuvers amidst unknown disturbances and boundary conditions.
    \item We introduce BLF-based constraints over an MPC controller to include obstacle avoidance and achieve tangible performance gain in trajectory tracking for the end-effector of the AM in comparison to prior MPC controllers \cite{NMPC}. 
    \item We exploit BLF to create a novel constraint for bounding the AM inside a defined boundary, contrary to it's collision avoidance utility.
    \item A disturbance resistivity term is introduced in the BLF for guaranteeing safety under bounded random disturbances. 
\end{enumerate}
    

\noindent The paper is structured as follows, Section II provides the dynamics model of the AM and BLF forward invariance constraint for collision avoidance. Section III provides the problem statement, while Section IV proposes a control architecture for the safe operation of AM. Section V discusses the simulation and benchmark results.

\section{PRELIMINARIES}

\subsection{Mathematical model of aerial manipulator}
In this section, we present the dynamics model of an AM \cite{2DOF design} in Newton-Euler Form. Here, the center of the UAV coincides with the center of gravity of the AM. We denote the Inertial frame $I = [I_x,I_y,I_z]$ with the centre at $O_I$, the Body frame fixed to the UAV is represented as $B = [B_x, B_y, B_z]$ centred at $O_B$. The manipulator link frames are denoted as $M_l = [M_{x_{l}}, M_{y_{l}}, M_{z_{l}}]$ centred at $O_l$. The position of the center of the UAV in the inertial frame is $\mathbf{p}_I = [x_I, y_I, z_I]^T$ and orientation in Euler angle convention is $\Phi = [\phi, \theta, \psi]^T$. Similarly, the manipulator joint angles are defined by $\Theta = [\theta_1,\theta_2,...,\theta_j]^T$ where $j$ is the number of joints, refer to Fig. \ref{fig1}. The acceleration vector due to gravitational forces is denoted by $\mathbf{g} = [0,0,g_z]$ where $g_z$=9.81 $m/s^2$. The mathematical models of UAV and manipulator are presented first, followed by the combined dynamics of AM.

\subsubsection{UAV Dynamics}
\par The translational dynamics of UAV is given in (\ref{eq1}).
\begin{equation}
\!
\begin{aligned}
\ddot{\mathbf{p}}_I &= \mathbf{g} + {^{I}R_B}\mathbf{T}_B/m_B 
\end{aligned}
\label{eq1}
\end{equation}
where $m_B$ is the mass of UAV, $T_B$ is the thrust vector acting on the UAV, defined in the body frame. $^{F_1}R_{F_2}$ denotes standard rotation matrix in 3D for transformation from frame $F_2$ to frame $F_1$ \cite{Rotation matrix}.


Denoting angular acceleration in the body frame as ${\mathbf{\omega}}_B$, the rotational dynamics is given in (\ref{eq2}).
\begin{equation}
\dot{\mathbf{\omega}}_B = \mathbf{I}_B^{-1}(\mathbf{\tau}_B - \dot{\mathbf{\omega}}_B \times \mathbf{I}_B\dot{\mathbf{\omega}}_B)
\label{eq2}
\end{equation}

where $\tau_B$ and $\mathbf{I}_B$ are respectively the torque acting on the UAV and inertia matrix, defined in the body frame.

\par Hence, combining (\ref{eq1}) and (\ref{eq2}) in matrix addition form, the UAV dynamics is shown below.

\begin{equation}
\label{eq3}
\resizebox{\hsize}{!}{$
\begin{bmatrix}
m_B\mathbf{I}_{3\times3} && \mathbf{0}_{3\times3} \\
\mathbf{0}_{3\times3} && \mathbf{I}_B \\
\end{bmatrix} 
\begin{bmatrix}
\ddot{\mathbf{p}}_I \\
\dot{\mathbf{\omega}}_B \\
\end{bmatrix} +
\begin{bmatrix}
\mathbf{0}_{3\times3} \\
\mathbf{\omega}_B \times \mathbf{I}_B\mathbf{\omega}_B \\
\end{bmatrix} 
=
\begin{bmatrix}
m_B\mathbf{g} + {^{I}R_B}T_B \\ 
\mathbf{\tau}_B \\
\end{bmatrix}$}
\end{equation} 

where $\mathbf{I}_{3\times3}$ is a 3$\times$3 identity matrix, $\mathbf{0}_{3\times3}$ is a $3\times3$ null matrix. 
\subsubsection{Dynamics of Floating Base Manipulator}

\par We derive the equation of motion for an individual link based on Newton-Euler formulation using standard Denavit-Hartenberg (DH) convention \cite{DH parameters}. Forward recursion starting from base link to the end-effector is used as
\begin{equation}
\begin{aligned}
\omega_{i} &= {^{i}\mathbf{R}_{i-1}}\omega_{i-1} + {^{i}\mathbf{R}_0}z_{i-1}\dot \theta_i\\
\alpha_{i} &= {^{i}\mathbf{R}_{i-1}}\alpha_{i-1} + {^{i}\mathbf{R}_0}z_{i-1}\ddot \theta_i + \omega_{i} \times {^{i}\mathbf{R}_0}z_{i-1}\dot \theta_{i}\\
\mathbf{a}_{e,i} &= {^{i}\mathbf{R}_{i-1}}\mathbf{a}_{e,i-1} + \dot \omega_i \times \mathbf{r}_{i,i+1} + \omega_i \times (\omega_i \times \mathbf{r}_{i,i+1})\\
\mathbf{a}_{c,i} &= {^{i}\mathbf{R}_{i-1}}\mathbf{a}_{e,i-1} + \dot \omega_i \times \mathbf{r}_{i,ci} + \omega_i \times (\omega_i \times \mathbf{r}_{i,ci})\\
\mathbf{v}_{e,i} &= {^{i}\mathbf{R}_{i-1}}\mathbf{v}_{e,i-1} + \mathbf{r}_{i-1,i} \times {^{i}\mathbf{R}_0}z_{i-1}\dot \theta_i\\
\mathbf{p}_{e,i} &= {^{i}\mathbf{R}_{i-1}}\mathbf{p}_{e,i-1} + \mathbf{r}_{i-1,i} \times {^{i}\mathbf{R}_0}z_{i-1}\theta_i
\end{aligned}
\label{eq4}
\end{equation}
where $\theta_i$ is the joint angle of joint i, $l_i$, $m_i$, $I_i$ are the link length, mass and inertia of link i respectively, $\mathbf{r}_{j,ci}$ is the vector from joint j to the CoM of link i, $\mathbf{r}_{j,l}$ is the vector from joint j to joint l, $\mathbf{a}_{c,i}$ is the acceleration of the center of mass of link i, $\omega_i$, $\alpha_i$ are the angular velocity and angular acceleration of frame i w.r.t. frame i, $z_{i-1}$ is the axis of rotation of joint i w.r.t. frame 0.

\par Backward recursion shows force and torque on link i as
\begin{equation}
\begin{aligned}
\mathbf{f}_i &= {^{i}\mathbf{R}_{i+1}}\mathbf{f}_{i+1} + m_i\mathbf{a}_{c,i} - m_i\mathbf{g}_i\\
\mathbf{\tau}_i &= {^{i}\mathbf{R}_{i+1}}\mathbf{\tau}_{i+1} - \mathbf{f}_i \times \mathbf{r}_{i,ci} + {^{i}\mathbf{R}_{i+1}}\mathbf{f}_{i+1} \times \mathbf{r}_{i+1,ci}\\& + I_i\alpha_i + \omega_i \times (I_i \times \omega_i)
\label{eq5}
\end{aligned}
\end{equation}

where $\mathbf{f}_i$, $\tau_i$ are the force and torque respectively exerted by link i-1 on link i with the terminal conditions as $\mathbf{f}_{n+1}$ = 0 and $\tau_{n+1}$ = 0.
\subsubsection{Dynamics of the Aerial Manipulator}
\par AM dynamics are written by adding the forces and torques from the manipulator given in (\ref{eq5}) to the UAV dynamics given in (\ref{eq4}) and utilizing the mass $m_{am}$ and moment of inertia $\mathbf{I}_{am}$ of the combined system. The coupled equation of motion for the UAV with the manipulator is given in (\ref{eq6}).
\begin{equation}
\resizebox{\hsize}{!}{$
\begin{bmatrix}
m_{am}\mathbf{I}_{3\times3} && \mathbf{0}_{3\times3}\\
\mathbf{0}_{3\times3} && \mathbf{I}_{am} \\
\end{bmatrix} 
\begin{bmatrix}
\ddot{\mathbf{p}}_I\\
\dot{\mathbf{\omega}}_B \\
\end{bmatrix} +
\begin{bmatrix}
\mathbf{0}_{3\times3}\\
\omega_B \times \mathbf{I}_{am}\omega_B \\
\end{bmatrix} =
\begin{bmatrix}
m_B\mathbf{g} + {^I\mathbf{R}_B}T_B+\mathbf{f}_0 \\ 
\mathbf{\tau}_B+\mathbf{\tau}_0\\
\end{bmatrix} $}
\label{eq6}
\end{equation}

\subsection{Barrier Lyapanov Function}

A BLF for the avoidance of a point obstacle by a UAV is presented here. The control affine form 
 of the UAV dynamics  given in (\ref{eq3})  is shown  below.

\begin{equation}
     \mathbf{\dot x} = f(\mathbf{x}) + g(\mathbf{x}) \mathbf{u}
     \label{eq7}
\end{equation}
where $f(\mathbf{x})$ and $g(\mathbf{x})$ are functions of the state $\mathbf{x}$ and $\mathbf{u}$ is the control input at time $t$.
 
$h(\mathbf{x})$ is a valid BLF if it is continuously differentiable and the following conditions given in (\ref{eq8}) are satisfied.
\begin{equation}
  \left\{\begin{array}{@{}l@{}}
    h(\mathbf{x}) \geq  0 , \ \forall \ \mathbf{x}  \in  \zeta\\
    h(\mathbf{x}) <  0 , \  \forall  \ \mathbf{x}  \notin  \zeta\\
  \end{array}\right.\,
  \label{eq8}
\end{equation}

where $\zeta$ denotes the set of all states in the safe region of operation and is defined as $\zeta = \{\mathbf{x} \in  \mathbb{R} ^{n} ~|~ h(\mathbf{x}) \geq 0\}$.
    

If initially, the UAV resides in the safe region and the condition $\dot h (\mathbf{x}) \geq 0$ implies that $\zeta$ is forward invariant. This ensures that $h(\mathbf{x})$ remains in the desired safe region. The forward invariance condition $\dot h (\mathbf{x}) \geq 0$ can be relaxed to $\dot h (\mathbf{x}) \geq - \gamma h^{z}(\mathbf{x})$ leading to asymptotic convergence of $h(x)$ to 0.  Considering the state-space model in (\ref{eq7}), the forward invariance condition can be written as given below.

\begin{equation}
    \frac{\partial h(\mathbf{x})}{\partial x} (f(\mathbf{x}) + g(\mathbf{x}) \mathbf{u}) + \gamma h^{z}(\mathbf{x}) \geq 0 
    \label{eq10}
\end{equation}
where $\gamma > 0 $ and $z > 0 $ are tunable parameters. BLF ($h_0$) for point obstacle avoidance satisfying the conditions given in (\ref{eq8}) and (\ref{eq10}) can be selected as shown below.

\begin{equation}
    h_0(\mathbf{x}) = \sqrt{2 \alpha _{max} (||\vec{\mathbf{p}}|| - d_s)} + \frac{\vec{\mathbf{p}}^T}{||\vec{\mathbf{p}}||} \vec{\mathbf{v}}
    \label{eq11}
\end{equation}

where $\alpha _{max}$ is the maximum acceleration attainable by the AM, $d_s$ is the desired safe distance, $\vec{\mathbf{p}}$ is the range vector from point obstacle and $\vec{\mathbf{v}}$ is the velocity of the robot at time $t$.  Differentiating $h_0(\mathbf{x})$ in (\ref{eq11}) and substituting in (\ref{eq10}) gives forward invariance condition on the controller as given in (\ref{eq12}).

\begin{equation}
\begin{aligned}
    \frac{\alpha_{max}~ \vec{\mathbf{v}}^T \vec{\mathbf{p}}}{\sqrt{2 \alpha _{max} (||\vec{\mathbf{p}}|| - d_s)}} - {\left (\frac{\vec{\mathbf{p}}^T}{||\vec{\mathbf{p}}||} \vec{\mathbf{v}} \right )}^{2} + ||\vec{\mathbf{v}}||^2  + \vec{\mathbf{p}}^T \mathbf{u} \\ +  \gamma h_0^z(\mathbf{x}) ||\vec{\mathbf{p}}|| ~\geq ~0
\end{aligned}
\label{eq12}
\end{equation}

\section{PROBLEM FORMULATION}

The primary aim of the paper is to design a controller to follow the desired trajectory ($\mathbf{p}^{d}$) for the end-effector of the AM i.e. to minimize at any time $t$ the trajectory error ($e_e(\mathbf{x})$)
\begin{equation}
     \underset{\mathbf{u}} {\min} \ e_{e}(\mathbf{x}) =  ||\mathbf{p}_{e}(\mathbf{x})- \mathbf{p}^{d}|| \  \forall \ t > 0
     \label{eq13}
\end{equation}
\vspace{-0.2em}
where $\mathbf{p}_{e}(\mathbf{x})$ is the position of the end-effector and $\mathbf{p}^{d}$ is the desired position of the end-effector at time $t$. This trajectory is defined for the end-effector, while the trajectory of the UAV is left to be determined by the controller, such as to reduce $\mathbf{e}_{e}(\mathbf{x})$ given in (\ref{eq13}).

\begin{figure}[] 
\centering
\includegraphics[width=0.38\textwidth]{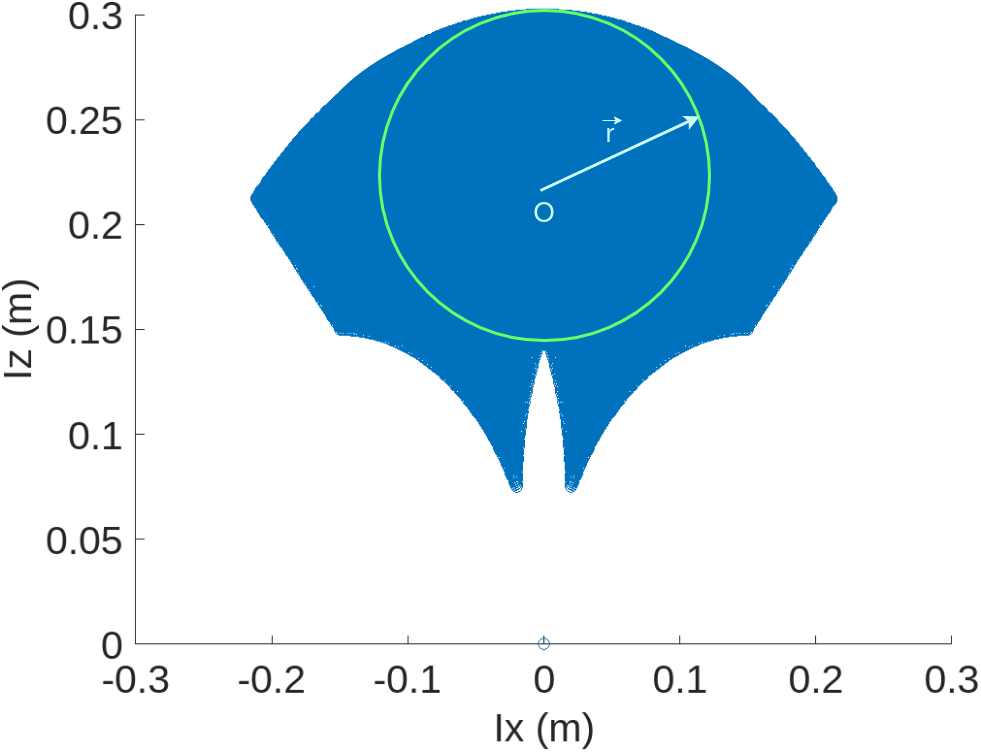}
\caption{\small{Inverse Workspace of a floating base manipulator wrt to a fixed end-effector point (0,0). This is a 2-D representation of a 3-D workspace. The 3-D workspace can be visualized by rotating 360\textdegree around line x = 0. The trajectory bound shown as a circle is a sphere $||p_r - O|| = r$ where $p_r$ is a point on the sphere, $O$ is the center of the sphere in 3D space and $r$ is the radius of the sphere.}}
\label{fig2}
\end{figure}

There are multiple ways to reach a desired point on the trajectory by placing the UAV in the inverse workspace of the floating base manipulator. This inverse workspace, Fig. \ref{fig2}, is defined by taking the inverse kinematics of the floating base manipulator considering it's end-effector to be fixed at the desired trajectory point. E.g., Fig. \ref{fig2} shows the inverse workspace of a 2R manipulator in which manipulator joint angles are constrained.

\begin{figure}[t]
\subfigure[] {\includegraphics[width=0.22\textwidth]{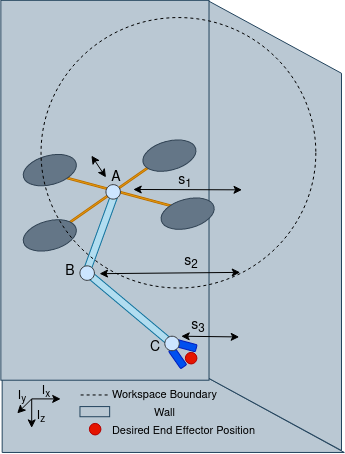}}~~~
\subfigure[] {\includegraphics[width=0.25\textwidth]{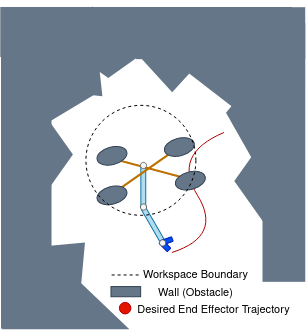}}
\centering
\caption{\small{\textbf{Setup for AM simulation:} (a) Barrier avoidance: Shows close operation of an AM with rectangular walls on the three sides with a known position. (b) Free space workspace tracking: Shows bounded maneuver of the AM in the desired workspace within an allowed free space of the operation (here, the obstacle position information is not available).} }
\label{fig3}
\end{figure}

Two cases are considered here, \textbf{Case I}, the location of the obstacle is known, and \textbf{Case II}, the allowed free space for operation is given without specifying the location of the obstacle. For \textbf{Case I} shown in Fig. \ref{fig3}(a), the end-effector of the AM has to follow the desired trajectory close to static obstacles. The condition for safety is given in (\ref{eq14}).
\begin{equation}
\begin{aligned}
    ||\vec{\mathbf{s}}_j ( \mathbf{x}_i )|| \geq s^{min}_j \ \forall \ j \in \{1,...,C\}\\
    \vec{\mathbf{s}}_j \times [a,b,c]^T = 0  
    \label{eq14}
\end{aligned}
\end{equation}
\vspace{-0.5em}

where $\vec{\mathbf{s}}_j$ is the perpendicular distance vector of the critical point $j$ from the wall and $s^{min}_j$ is the minimum safety distance from the wall. The number of critical points is denoted by $C$. 
In Fig. \ref{fig3}(a), critical points on the AM are (A) the base of the manipulator with safety radius ${s}_1$, (B) joint 2 of the manipulator with safety radius ${s}_2$ and (C) end-effector of the manipulator with safety radius ${s}_3$. Equation of the wall is given by $ax + by + cz + d = 0$ where $a$,$b$,$c$,$d$ are parameters of the plane and $x$, $y$, $z$ are points on the plane.


For \textbf{Case II}, Fig. \ref{fig3}(b), a safe free space of operation around the trajectory is provided, and we provide a guarantee that the AM will stay in that safe bounded region. We choose a spherical differentiable part Fig. \ref{fig2} of the non-differentiable inverse workspace with radius $r$. The condition for bounding in the desired workspace is given in (\ref{eq15}).

\begin{equation}
\begin{aligned}
    ||\vec{\mathbf{d}} (\mathbf{x} )|| \leq r,\,\,\,
    \vec{\mathbf{d}} (\mathbf{x}_i) = \mathbf{p}_{I}(\mathbf{x} )- (\mathbf{p}^{d} - d_{iw})
    \label{eq15}
\end{aligned}
\end{equation}
where $r$ is the radius of the desired workspace and $d_{iw}$ is the deviation of the workspace center from the desired trajectory point. As the bound for the allowable free space is within the inverse workspace, the end-effector would be able to reach the desired point in at least one orientation of the AM. 



\begin{figure*}[ht]
\includegraphics[scale=0.44]{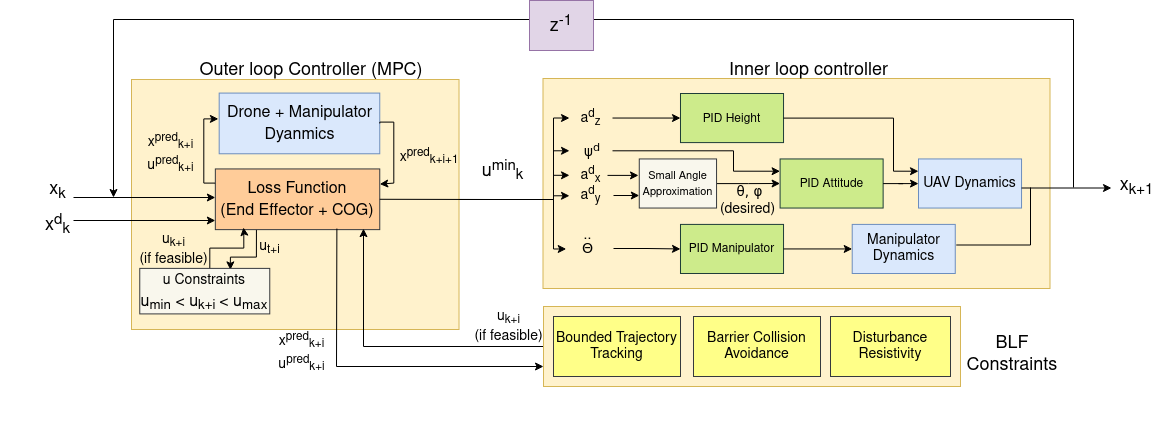}
\caption{\small{\textbf{Control Architecture of the AM.} Outer loop controller denotes the MPC optimization problem while the inner loop controller denotes the on-board PID control of the AM. Constraints on the optimizer are denoted by input constraints ($\mathbf{u}$) an Constraints.}}
\label{fig4}
\end{figure*}

\section{PROPOSED CONTROLLER}
The architecture of the proposed controller is shown in Fig. \ref{fig4}. It consists of an outer loop MPC  with BLF constraints. The inner loop is driven by a conventional PID controller. The outer loop provides the desired acceleration and yaw angle references to the inner loop controller.
\subsection{MPC (Outer loop control)} 
The state vector of the AM is defined as $\mathbf{x} = [\mathbf{p}_I, \mathbf{\dot{p}}_I, \psi, \dot\psi, \Theta, \dot\Theta]$, where $\mathbf{p}_I$ and $\mathbf{\dot{p}}_I$ is the position and velocity of the center of mass of the UAV in inertial frame respectively, $\psi$ and $\dot\psi$ is the yaw and yaw rate, and $\Theta = [\theta_1, \theta_2]$ and $\dot\Theta = [\dot\theta_1, \dot\theta_2]$ is the joint angle and angular rate of the manipulator's joints.
The control input for MPC is $\mathbf{u}= [\mathbf{a}_d, \ddot\psi, \ddot\Theta_d]$ where $\mathbf{a}_d = [a_x, a_y, a_z]$ is the acceleration of the UAV center in the inertial frame, $\ddot\psi$ is the yaw angular acceleration and $\ddot\Theta$ is the angular acceleration of manipulator's joints. The state space model used within the MPC is formulated as given in (\ref{eq15}).

\begin{equation}
\begin{aligned} 
\mathbf{\dot x} = \mathbf{A}\mathbf{x} + \mathbf{B}\mathbf{u} \\
\end{aligned}
\label{eq15}
\end{equation}

\begin{equation}
\mathbf{A} = \begin{bmatrix}
0_{3x3} && I_{3x3} && 0_{3x6}\\
0_{3x3} && 0_{3x3} && 0_{3x6}\\
0_{1x7} && 1 && 0_{1x4}\\
0_{1x3} && 0_{1x3} && 0_{1x6}\\
0_{2x3} && 0_{2x7} && I_{2x2}\\
0_{2x3} && 0_{2x3} && 0_{2x6}\\
\end{bmatrix} 
\mathbf{B} = \begin{bmatrix}
0_{3x3} && 0 && 0_{3x2}\\
I_{3x3} && 0 && 0_{3x2}\\
0_{1x3} && 0 && 0_{1x2}\\
0_{1x3} && 1 && 0_{1x2}\\
0_{1x3} && 0 && 0_{1x2}\\
0_{1x3} && 0 && I_{1x2}
\end{bmatrix}\\\\
\label{eq16}
\end{equation} 

In (\ref{eq16}), ${I}$ is a diagonal matrix such that, $I(i,j) = 1$, $\forall$ $i=j$ and $I(i,j) = 0$, $\forall$ $i \neq j$ and ${0}$ is zero matrix such that, $0(i,j) = 0$, $\forall$ $i,j$ where $i,j\in\mathbb{N}$.

For Aerial Manipulator dynamics with state variables as $\mathbf{x}$ and control variables as $\mathbf{u}$,
the optimal control problems at every time instant $t_k$ where $t_k = kT$ and $T$ is a time step is given in (\ref{eq17}).
\begin{subequations}
\label{eq:Cost_function}
\begin{align}
\underset{\mathbf{u}} {\min} \ l( \mathbf{x_k,u_k}, t_{k})\\
\textrm{s.t.} \  \mathbf{x}_{k+1} = A \mathbf{x}_{k} + B \mathbf{u}_{k}\\
\mathbf{u}_{min} \leq \mathbf{u}_k \leq \mathbf{u}_{max}\\
\mathbf{x}_{min} \leq \mathbf{x}_k \leq \mathbf{x}_{max} 
\end{align}
\label{eq17}
\end{subequations}

where $\mathbf{u} = [{\mathbf{u}_k}^\top, {\mathbf{u}_{k+1}}^\top, ...., {\mathbf{u}_{k+N-1}}^\top]^\top $ denotes the vector of control variables and $f$ is the dynamic model of the system. $\mathbf{u}_{min}$ and $\mathbf{u}_{max}$ represent the bounds on $\mathbf{u}$ and $\mathbf{x}_{min}$ and $\mathbf{x}_{max}$ represent the bounds on $\mathbf{x}$. $\mathbf{x}_{k}$ denotes the state of the system at $k^{th}$ time step, similarly $\mathbf{u}_{k}$ denotes the control input of the system at $k^{th}$ time step. The cost $l$ is the summation of multiple cost functions $l = \sum_{i = 1}^{N_l} l_i$ explained in the section \textit{Weighing strategy}.

\subsubsection{Weighing strategy}

The $N_{l}$ costs are chosen to follow the trajectory and enhance stability. These are defined in (\ref{eq18}), (\ref{eq20}) and (\ref{eq21}).

\paragraph{Tracking error for the End-Effector}

The primary task is the tracking of the end effector over a given trajectory. This is achieved by penalizing the difference between the current and desired position as given in (\ref{eq18}) 

\begin{equation}
\begin{aligned}
l_{1} = \ \sum_{i = 0}^{n-1} (  \left|\left|   \mathbf{e}_{e}(\mathbf{x}_{k+i})   \right|\right|^2_{W_{1}} ) + \left|\left|  \mathbf{e}_{e}( \mathbf{x}_{k+N})  \right|\right|^2_{W_{s_{1}}}
\end{aligned}
\label{eq18}
 \end{equation}
 
 where $\mathbf{e}_{e}(\mathbf{x}_{k+i}) =  \ \mathbf{p}_{e}(\mathbf{x}_{k+i} )- \mathbf{p}^{d}_{k+i}$.
 
     

is the end-effector tracking error calculated at each time step of the prediction horizon. To decrease error from the desired trajectory, we penalize higher velocities of the manipulator end-effector as given in (\ref{eq20}).

\begin{equation}
\begin{aligned}
l_{2} = \ \sum_{i = 0}^{n-1} (  \left|\left|   \mathbf{v}_{e}(\mathbf{x}_{k+i})   \right|\right|^2_{W_{2}} ) + \left|\left|  \mathbf{v}_{e}( \mathbf{x}_{k+N})  \right|\right|^2_{W_{s_{2}}}
\end{aligned}
\label{eq20}
\end{equation} 

where $\mathbf{v}_{e}$ is the velocity of the End-Effector of the Manipulator.

\paragraph{COG Allignment Error}

As the manipulator moves in the $\mathbf{B_x B_z}$ plane of the UAV (refer to Fig. \ref{fig1}) while the UAV changes its attitude $(\theta, \phi, \psi)$, the Center of Gravity of the system moves in the $\mathbf{B_x B_y}$ direction of the UAV due to which undesirable torques appear which destabilize the UAV. The following cost given in (\ref{eq21}) is introduced considering this factor.
\begin{equation}
\begin{aligned}
l_{3} = \ \sum_{i = 0}^{n-1} (  \left|\left|   \mathbf{p}_{{G}_{XY}}(\mathbf{x}_{k+i})   \right|\right|^2_{W_{3}} ) + \left|\left|  \mathbf{p}_{{G}_{XY}}( \mathbf{x}_{k+N})  \right|\right|^2_{W_{s_{3}}}
\end{aligned}
\label{eq21}
\end{equation} 

where $\mathbf{p}_{{G}_{XY}}$ is the Center of Gravity of the Manipulator in the $\mathbf{B_xB_y}$ 
plane. \\

MPC to reach desired points using this approach was used in \cite{NMPC}, and we would call this technique \textbf{Naive MPC}. 

\subsection{PID (Inner Loop Control)}

From the control input ($\mathbf{u}_{k}^{min}$) obtained from the outer loop controller, the desired values of $\theta$, $\phi$ are estimated using small angle approximation (refer to Fig. \ref{fig4}) and along with $\psi_d$ are tracked by the PID controller namely PID Attitude. Similarly, PID Height regulates the height, and PID Manipulator angle controls the joint angles $\theta_1$ and $\theta_2$ \cite{Adaptive control}.

\subsection{Safe operation near known barriers}

For \textbf{Case I}, the priority for safety is the highest. The AM has to avoid collisions for both the UAV and the manipulator simultaneously while tracking the desired trajectory. 
The critical points are chosen such that if we can guarantee collision avoidance for these points, the entire system can safely perform desired maneuvers with collision. 
BLF used for collision avoidance along the radial direction of the Wall is given in (\ref{eq22}). 

\begin{equation}
\noindent h_1(\mathbf{x}_k) = \sqrt{\frac{2 \vec{\mathbf{s}}_j(\mathbf{x}_{k+i})^T \alpha_{max}}{||\vec{\mathbf{s}}_j(\mathbf{x}_{k+i})||}  (||\vec{\mathbf{s}}_j(\mathbf{x}_{k+i})|| - s^{min}_j)} + \frac{\vec{\mathbf{s}}_j(\mathbf{x}_{k+i})^T}{||\vec{\mathbf{s}}_j(\mathbf{x}_{k+i})||} \vec{\mathbf{v}}_{k+i} 
\label{eq22}
\end{equation}



\subsection{Bounded Trajectory Tracking }


\begin{figure}[t]
\includegraphics[scale=0.4]{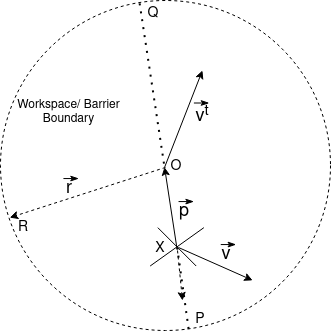}
\centering
\caption{\small{Desired workspace for the UAV (refer to the circle in Fig. \ref{fig2}). X denotes the Center of the UAV, $\vec{p}$ denotes vector from X to O, $\vec{v}$ denotes instantaneous velocity of X and $\vec{v}_k$ denotes instantaneous velocity of desired trajectory of the end-effector}}
\label{fig5}
\end{figure}

\noindent For \textbf{Case II}, one of our major contributions in this section is the usage of BLF for the bounding of the UAV in the desired workspace shown in Fig. \ref{fig5}. We exploit the property that a particle can exit a sphere only through its motion in the radially outwards direction. Hence, the UAV can only escape the boundary if it is provided with a high velocity in the radial direction (along $\vec{\mathbf{p}}$ or $-\vec{\mathbf{p}}$ ). 
Due to this restricted movement in the radial directions on both sides, the UAV does not leave the workspace at that particular instant $k$. BLF ($h_{2_a}$) along $\vec{\mathbf{p}}$ is given in (\ref{eq23}) and a similar BLF ($h_{2_b}$) along $-\vec{\mathbf{p}}$ can be found by replacing $\vec{\mathbf{p}}$ with $-\vec{\mathbf{p}}$.

\begin{equation}
h_{2_{a}}(\mathbf{x}_k) = \sqrt{\frac{2 (\vec{\mathbf{p}}_{k+i})^T \alpha_{max}}{||\vec{\mathbf{p}}_{k+i}||}  (||\vec{\mathbf{p}}_{k+i}|| - r)} + \frac{\vec{\mathbf{p}}_{k+i}^T}{||\vec{\mathbf{p}}_{k+i}||} (\vec{\mathbf{v}}_{k+i} - \vec{\mathbf{v}}^t_{k+i})
\label{eq23}
\end{equation}


\subsection{BLF with Disturbance Rejection}

Another contribution of this paper involves handling unseen disturbances, such as constant wind or impulses acting at random intervals in the framework of BLF. The condition in (\ref{eq10}) need not be satisfied for the given $h(\mathbf{x})$ in the presence of unmeasured disturbances. If the AM encounters a sudden disturbance, it might result in a collision with obstacles. The condition  $\dot h (\mathbf{x}) \geq - \gamma h^{z}(\mathbf{x})$ is tightened while maintaining relaxation for smooth operation as given in (\ref{eq25}) with $\lambda>0$.
\begin{equation}
    \dot{h}_i(\mathbf{x}) + \gamma({h}^z_i(\mathbf{x}) - \lambda) \geq 0  
    \label{eq25}
\end{equation}








\noindent The BLF functions $h_1$, $h_{2_a}$ and $h_{2_b}$ are put in (\ref{eq25}) and are added to the MPC optimizer as constraints. The resultant controller is termed as \textbf{MPC-BLF}. To the best of the author's knowledge, there have been no attempts to handle disturbance forces or torques for an AM using a BLF.  

\section{SIMULATION RESULTS}

The algorithm is implemented on Python 3 on an Intel® Core™ i7-8550U CPU PC running at 1.80 GHz. We use the 'SLSQP' method from scipy \cite{scipy} as the non-linear optimizer for MPC. The AM used for simulation is a mathematical replica of a laboratory scale AM shown in Fig. \ref{fig1} with the following specifications given in Table \ref{table1}.

\begin{table}[h]
\label{table_example}
\centering
\resizebox{\columnwidth}{!}{%
\begin{tabular}{|>{\centering}p{0.17\textwidth}|>{\centering\arraybackslash}p{0.26\textwidth}|}
\hline
\textbf{Parameter} &  \textbf{Value} \\
\hline
Mass &   $3.5~kg$ \\
\hline
Arm length & $0.3~m$\\
\hline
Propeller Diameter & $0.33~m$ \\
\hline
Moment of Inertia - UAV&  $I_x = 0.3~kg~m^2$, $I_y = 0.3~kg~m^2$, $I_z = 0.6~kg~m^2$\\
\hline
Manipulator length & $link 1 = 0.15~m$, $link 2 = 0.15~m$  \\

\hline
Moment of Inertia - Manipulator&  $I_1 = 4.256\times10^{-5}~kg~m^2$, $I_2 = 8.321\times10^{-5}~kg~m^2$ \\
\hline

Moment of Inertia - Manipulator&  $I_1 = 4.256\times10^{-5}~kg~m^2$, $I_2 = 8.321\times10^{-5}~kg~m^2$ \\
\hline
UAV attitude constraints & $|\theta| \leq \pi/10~ rad$ , $|\phi| \leq \pi/10~ rad$\\ 
\hline
Manipulator joint angle constraints & $|\theta_1| \leq \pi/3~ rad $, $|\theta_1 + \theta_2| \leq \pi/2~ rad$ \\
\hline
\end{tabular}}
\caption{Specifications of the Aerial Manipulator}
\label{table1}

\end{table}

We perform simulations separately for two cases.

\textbf{Case I}: Avoiding Wall on three sides of the AM (The position of the obstacle is known).

\textbf{Case II}: Free space for the operation of AM is given (The position of obstacles is unknown, a  desired spherical workspace  for the UAV is created.) 

For both of the simulations, the task is to follow a desired trajectory by the end-effector. A uniformly random disturbance with an amplitude of $d_m=0.8\,m/s^2$ is used to evaluate the performance in the presence of external disturbances.

\subsection{Modified Naive MPC for comparison}

The Naive MPC is modified as given below  to compare the performance with the proposed controller.

\subsubsection{\textbf{Hard Constraint (MPC-HC)}} 

\noindent For \textbf{Case I}, a hard constraint on the critical points to avoid collision with the Wall is enforced as in (\ref{eq14}) $\forall~i~\in [0,n]$.
For \textbf{Case II}, a hard constraint on the relative position of UAV center (Fig. \ref{fig5}) as in (\ref{eq15}) $\forall~i~\in [0,n]$ is added. Similar to \textbf{Case I}, this is the intuitive constraint for bounding.
These additional constraints are put on the MPC optimizer in (\ref{eq17}).

\subsubsection{\textbf{Soft Constraint (MPC-SC)}} 

\noindent For \textbf{Case I}, a cost function for safe operation near walls is given in (\ref{eq29}). This cost increases as the Wall is approached, penalizing the critical points going near  the Wall.
\begin{equation}
\begin{aligned}
l_{4} = \ \sum_{i = 0}^{n-1}   \frac{1}{(||\vec{\mathbf{p}}_j ( \mathbf{x}_{k+i} )|| - s_j)^2_{W_{4}}}  +    \frac{1}{(||\vec{\mathbf{p}}_j ( \mathbf{x}_{k+n} )|| - s_j)^2_{W_{s_{4}}}}
\end{aligned}
\label{eq29}
 \end{equation}

For \textbf{Case II}, a cost function that penalizes the UAV movement outside the workspace is given in (\ref{eq30}) and is added to the MPC cost.
\begin{equation}
\begin{aligned}
l_{5} = \ \sum_{i = 0}^{n-1}   \left|\left|   \mathbf{p}(\mathbf{x}_{k+i})   \right|\right|^2_{W_{5}}  + \left|\left|  \mathbf{p}( \mathbf{x}_{k+n})  \right|\right|^2_{W_{s_{5}}} \ \forall \ |\mathbf{p}(\mathbf{x}_{k+i})| \geq r
\end{aligned}
\label{eq30}
 \end{equation}
Parameters and weights for the different MPC versions are given in Table \ref{table2}.

\begin{table}[h]
\begin{center}
{\begin{tabular}{|>{\centering}p{0.15\textwidth}|>{\centering\arraybackslash}p{0.28\textwidth}|}
\hline
\textbf{Parameter} &  \textbf{Value} \\
\hline
MPC Weights &   $w_1 = 10\times \mathbf{I}_{3\times3}$, $w_{s_1} = 50\times \mathbf{I}_{3\times3}$, $w_2 = 2\times \mathbf{I}_{3\times3}$, $w_{s_2} = 10\times \mathbf{I}_{3\times3}$, $w_3 = \mathbf{I}_{3\times3}$, $w_{s_3} = 5\times \mathbf{I}_{3\times3}$, $w_4 = 5\times \mathbf{I}_{3\times3}$, $w_{s_4} = 20\times \mathbf{I}_{3\times3}$, $w_5 = 5\times \mathbf{I}_{3\times3}$, $w_{s_5} = 20\times \mathbf{I}_{3\times3}$ \\
\hline
$s_j^{min}$ & $0.1$ m\\
\hline
$u_k$ Initialization & $\mathbf{0}_{1 \times 6n}$ \\
\hline
$d_{iw}$ &  $0.225$ $m$\\
\hline
$\gamma$ & $3$  \\
\hline
Sampling time ($t_s$) & $0.1$ s \\
\hline
Total time ($t$) & $120$ s \\
\hline
Max disturbance amplitude & $d_m  = 0.8\, m/s^2$ \\
\hline

\hline
\end{tabular}}
\caption{Weights and Parameters for MPC}
\label{table2}
\end{center}
\end{table}
\subsection{Metrics for performance comparison}


\noindent The performance of the controller is evaluated by the following metrics for  $N$ time steps ($N = t_f/t_s$). Here, $t_f$ is the final time, and $t_s$ is the sampling time in seconds.

\begin{itemize}
    \item Maneuver completion without collision or escaping the desired workspace 
    \item Manipulator end-effector root mean square error from the desired trajectory, TE = $\sqrt{\frac{1}{N} \sum_{k = 0}^{N-1} (\mathbf{e}_{e}(\mathbf{x}_{k}))^2}$
    \item Control effort, $c_e = \sum_{k = 0}^{N-1}{||\mathbf{u}_k||^2}$ 
    \item Control Smoothness, $c_s = \sum_{k = 0}^{N-1}{|\Delta\mathbf{u}_k|}$ 
\end{itemize}

\subsection{Performance comparison for \textbf{Case II}}
Selection of the prediction horizon ($n$) and the parameter $\lambda$ determines the computational load and disturbance rejection capabilities, respectively.
An analysis was conducted by varying n and $\lambda$ and the results are given in Table \ref{table3}. We chose n = 5 and $\lambda = 5$ as it shows low TE while maintaining the workspace bound. These values are used for all subsequent simulations. $n = 10$ does not show significant improvement (<5\%) in TE compared to $n=5$ and is rejected because of a very high computational time compared to $n=5$ (>600\%). $\lambda = 5$ shows high disturbance resistivity given its low TE and low $c_s$. $T_C$ denotes the computational time for one step.

\begin{table}[htbp]
\centering
\resizebox{\columnwidth}{!}{%
{\begin{tabular}{|>{\centering}p{0.085\textwidth}|>{\centering}p{0.047\textwidth}|>{\centering}p{0.047\textwidth}|>{\centering}p{0.047\textwidth}|>{\centering}p{0.047\textwidth}|>{\centering}p{0.047\textwidth}|>{\centering\arraybackslash}p{0.047\textwidth}|}
\hline
Parameter & \multicolumn{2}{c|}{n = 1} & \multicolumn{2}{c|}{n = 5} & \multicolumn{2}{c|}{n = 10}\\
 
\cline{2-3} \cline{4-5} \cline{6-7} 
        &  $\lambda = 1$ & $\lambda = 5$ & $\lambda = 1$ & $\lambda = 5$
	& $\lambda = 1$ & $\lambda = 5$  \\    \hline

\hline
$T_C$ (s) & 0.0313  & 0.0309  & 0.1712 & 0.1603 & 1.370 & 1.232 \\
\hline
TE (m) & 0.1943 & 0.1839 & 0.0738 & 0.0658 & 0.0701 & 0.0623\\
\hline
$c_s$  & 0.1593 & 0.1603 & 0.0746 & 0.0772 & 0.0716 & 0.0694 \\
\hline
$c_e$ & 0.1432 & 0.1248 & 0.0801 & 0.0788 & 0.0939 & 0.0942 \\
\hline
\end{tabular}}}
\caption {Ablation showing the effect of parameters on the proposed MPC-BLF method (units mentioned in brackets)}
\label{table3}
\end{table}

\begin{figure}[htbp]
    \subfigure[] {\includegraphics[width=0.93\columnwidth]{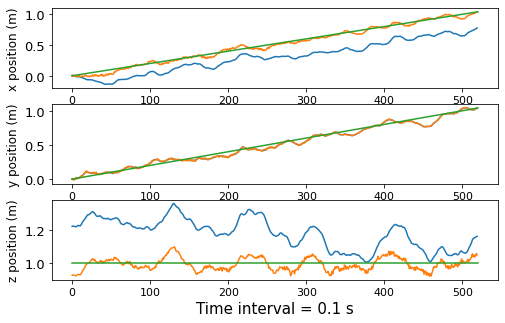}}\vspace{-1em}
    \subfigure[] {\includegraphics[width=0.93\columnwidth]{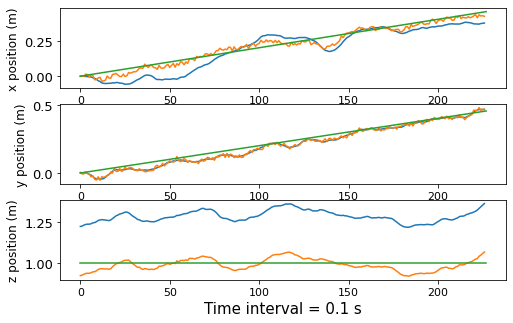}}\vspace{-1em}
    \subfigure[] {\includegraphics[width=0.93\columnwidth]{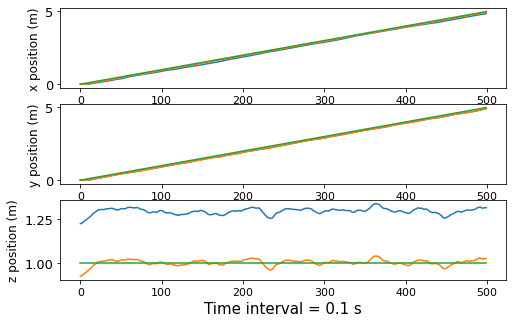}}\vspace{-1em}
    \subfigure[] {\includegraphics[width=0.93\columnwidth]{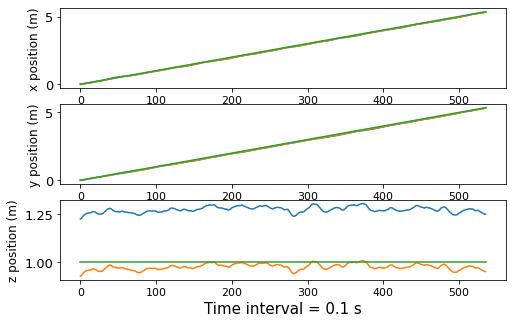}}\vspace{-1em}
    \caption{\textbf{AM position for Case II with disturbances:} (a) Naive MPC , (b) MPC-HC, (c) MPC-SC, (d) MPC- BLF. <green> -> Desired trajectory, <blue> -> UAV Center, <orange> -> End-effector position}
    \label{fig6}
\end{figure}

\begin{figure}[htbp]
    \subfigure[] {\includegraphics[width=0.48\columnwidth]{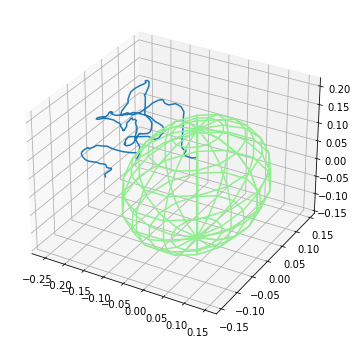}}\vspace{-0.4em}
    \subfigure[] {\includegraphics[width=0.48\columnwidth]{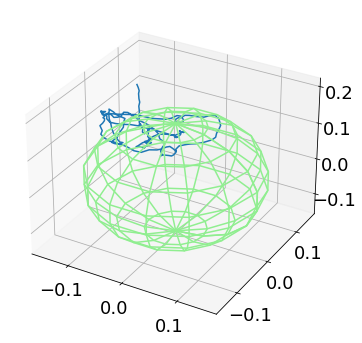}}\vspace{-0.4em}
    \subfigure[] {\includegraphics[width=0.48\columnwidth]{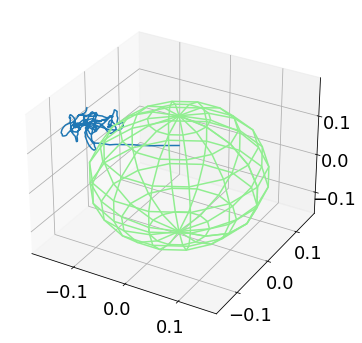}}\vspace{-0.4em}
    \subfigure[] {\includegraphics[width=0.48\columnwidth]{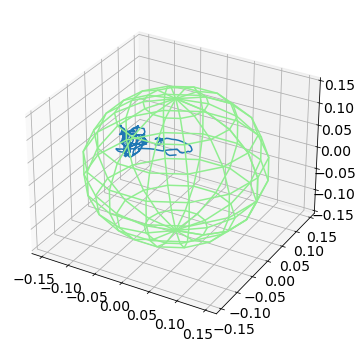}}\vspace{-0.4em}
    \caption{{Relative position of the UAV center w.r.t. the workspace bound (\textbf{Case II}) during the maneuvers in Fig. \ref{fig6}: } (a) Naive MPC , (b) MPC-HC, (c) MPC-SC, (d) MPC- BLF. <lightgreen> -> desired workspace boundary, <blue> -> UAV center. UAV Center should not move out of the workspace boundary.}
    \label{fig7}
\end{figure}

 The results given in Table \ref{table4} show that none among the Naive MPC, MPC-HC, and MPC-SC is able to restrict the UAV in the desired workspace, either in the presence or absence of external disturbances. MPC-HC has no penalization for the velocity of the UAV; hence when a large control input is provided, it exits the bounds. MPC-SC has to trade-off between TE and bounding cost, hence compromising on one of the factors. The proposed MPC-BLF is able to restrict the UAV within the desired workspace with the lowest control effort, highest control smoothness, and lowest tracking error. The inference is more evident in the presence of external disturbances. The trajectory of the AM is shown in Fig. \ref{fig6} for all four methods in the presence of external disturbances. The proposed MPC-BLF method is the only one to confine the UAV within the safe boundary (Fig. \ref{fig7}(d)).


\begin{figure}[htbp]
    \subfigure[] {\includegraphics[width=0.9\columnwidth]{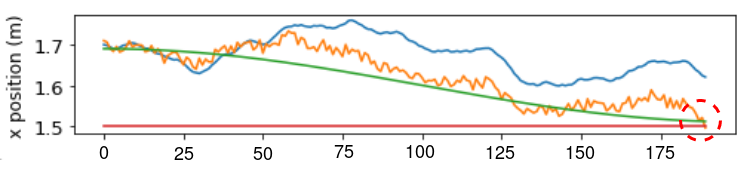}}\vspace{-1em}
    \subfigure[] {\includegraphics[width=0.89\columnwidth]{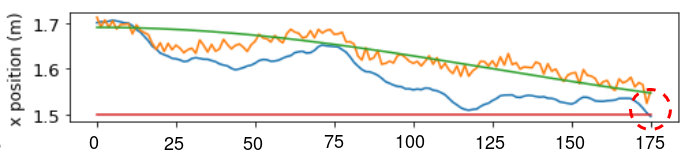}}\vspace{-1em}
    \subfigure[] {\includegraphics[width=0.9\columnwidth]{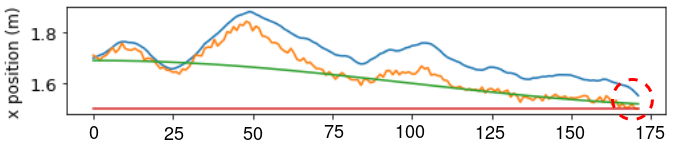}}\vspace{-1em}
    \subfigure[] {\includegraphics[width=0.9\columnwidth]{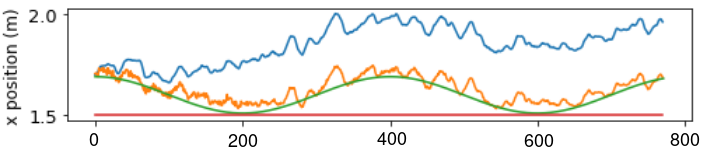}}\vspace{-1em}
    \caption{\textbf{AM position for Case I with disturbances in $I_x$ direction:} Red circle depicts collision with any wall. (a) Naive MPC: End-effector collides with the Wall (b) MPC - HC : UAV collides with Wall (c) MPC - SC : End-effector collides with the Wall (d) MPC-BLF: Maneuver completion.
    <green> -> Desired trajectory, <blue> -> UAV Center, <orange> -> End-effector position, <Red> -> Wall.}
    \label{fig8}
\end{figure}

\subsection{Performance comparison for \textbf{Case I}}

As shown in Fig. \ref{fig8}, \textbf{Case I} shows similar behaviour as \textbf{Case II}. Fig. \ref{fig8} shows trajectory tracking in only $I_x$, a similar result is obtained in $I_y$, $I_z$. As shown in Table \ref{table4}, MPC-SC shows a very high TE compared to MPC-BLF even in the absence of disturbances. MPC-BLF is able to avoid the walls contrary to any other method in the presence of disturbances.



\begin{table*}[h]
\centering

\resizebox{\hsize}{!}{%
\fbox{\begin{tabular}{|p{0.09\textwidth}|>{\centering}p{0.09\textwidth}||>{\centering}p{0.085\textwidth}|>{\centering}p{0.085\textwidth}||>{\centering}p{0.085\textwidth}|>{\centering}p{0.085\textwidth}||>{\centering}p{0.085\textwidth}|>{\centering}p{0.085\textwidth}||>{\centering}p{0.085\textwidth}|>{\centering\arraybackslash}p{0.085\textwidth}|}

\hline
  &  & \multicolumn{2}{c||}{Naive MPC}& \multicolumn{2}{c||}{MPC - HC}&  
 \multicolumn{2}{c||}{MPC- SC} &   \multicolumn{2}{c|}{MPC- BLF}\\
\cline{3-4} \cline{5-6} \cline{7-8} \cline{9-10}
        &  &   \textbf{Case I}& \textbf{Case II} &  \textbf{Case I} & \textbf{Case II}
	&  \textbf{Case I} & \textbf{Case II}  & \textbf{Case I} & \textbf{Case II} \\    
 \hline

\multirow{5}{*}{Without Disturbance} &  & $\times$ & $\times$ & $\times$ &$\times$ & $\checkmark$ & $\times$ & $\checkmark$ & $\checkmark$ \\
\cline {2-10}
& TE (m) '$\downarrow$' & $0.02432*$ & $0.02429$ & $0.02657*$ & $0.02386$ & $0.04482$ & $0.09731$ & $\mathbf{0.02629}$ & $\mathbf{0.04750}$  \\
\cline {2-10}
& $c_s$ '$\downarrow$'& $0.00557*$ & $0.00612$ & $0.02880*$ & $0.04981$ & $\mathbf{0.00443}$ & $0.00698$ & $0.00732$ & $\mathbf{0.03212}$ \\
\cline {2-10}
& $c_e$ '$\downarrow$'& $0.01427*$ & $0.01489$ & $0.04787*$ & $0.06572$ & $0.02326$ & $0.02229$ & $\mathbf{0.022771}$ & $\mathbf{0.04531}$ \\
\cline {2-10}
& & & & & & & & & \\
\hline 
 
\multirow{5}{*}{With Disturbance} &   & $\times$ & $\times$ & $\times$ & $\times$ & $\times$ & $\times$ & $\checkmark$ & $\checkmark$ \\
\cline {2-10}
& TE (m) '$\downarrow$'& $0.07366*$ & $0.06735$ & $0.05430*$ &  $0.05490*$ & $0.07708*$ & $0.11384$ & $\mathbf{0.07164}$ & $\mathbf{0.06589}$ \\
\cline {2-10}
& $c_s$ '$\downarrow$'& $0.08933*$ & $0.08942$ & $0.10546*$ & $0.11304*$ & $0.08933*$ & $0.07915$ & $\mathbf{0.15592}$ & $\mathbf{0.07720}$ \\
\cline {2-10}
& $c_e$ '$\downarrow$'& $0.17720*$ & $0.20924$ & $0.18921*$ & $0.18301*$ & $0.21039*$ & $0.08328$ & $\mathbf{0.23040}$ & $\mathbf{0.07887}$ \\
\cline {2-10}
& & & & & & & & & \\
\hline 
\end{tabular}}}

\caption { \textbf{Algorithm Benchmarking for simulations: } For Wall Avoidance (\textbf{Case I}), '$\times$' signifies that the AM collided with the wall. For workspace bound (\textbf{Case II}), '$\times$' signifies that the UAV was unable to maintain its position inside the desired workspace. '*' denotes that the maneuver was incomplete due to collision with the Wall or the inability of the MPC optimizer to find a valid control input to satisfy safe operation conditions. '$\downarrow$' denotes that a lower value of the performance metric is desired.}
\label{table4}
\end{table*}

\section{CONCLUSION}
This paper presented a BLF-based Model predictive controller with a primary objective of safe operation in the proximity of static objects.
Our approach shows how BLF, formulated for barrier avoidance and free space tracking objectives, shows robust behavior for the safe operation of an Aerial Manipulator in the presence of external disturbances. 
A state-of-the-art MPC-based method is modified using two types of constraints and compared with the proposed method, which shows significant improvement in two cases i.e with and without external disturbances. 
This is validated in simulation using parameters from a real laboratory scale AM.




\end{document}